\documentclass[12pt]{article}
\usepackage[french]{babel}
 \usepackage{url}
 \usepackage{amsmath}
 \usepackage{amsfonts}
 \usepackage[section]{algorithm}
\floatname{algorithm}{Algorithme}
\usepackage{algorithmicx}
\usepackage{color,colortbl,graphics}
\usepackage[utf8]{inputenc}
\begin{document}


\begin{center}
{\Large
	{\sc  S\'election de variables par le GLM-Lasso pour la pr\'ediction du risque palustre.}
}
\bigskip

 Bienvenue  Kouway\`e $^{1,2}$,  No\"el Fonton $^{2}$  \& Fabrice Rossi $^{3}$ 
\bigskip

{\it
$^{1}$ Uiversit\'e paris 1, 90 rue de Tolbiac, kouwaye2000@yahoo.fr 
 
$^{2}$ Universit\'e d'Abomey-Calavi, CIPMA   072 BP 50 Cotonou, hnfonton@gmail.com

$^{3}$ Uiversit\'e paris 1, 90 rue de Tolbiac, Fabrice.Rossi@univ-paris1.fr 
}
\end{center}
\bigskip

{\bf R\'esum\'e.} Nous \'etudions dans ce travail une m\'ethode de  s\'election 
de variables bas\'ee sur le Lasso dans le contexte \'epid\'emiologique.  L'un des objectifs est de construire automatiquement un mod\`ele pr\'edictif en limitant le 
recours aux experts m\'edicaux qui  op\`erent des pr\'etraitements sur les donn\'ees collect\'ees. Ces pr\'etraitements  consistent 
entre autres \`a recoder certaines variables en classe et \`a choisir  manuellement certaines interactions en se basant 
sur la connaissance des donn\'ees.  L'approche propos\'ee utilise toutes les variables explicatives sans traitement et  g\'en\`ere automatiquement toutes 
les interactions entre les variables, ce qui nous conduit en grande dimension. Nous utilisons le Lasso qui est une 
m\'ethode robuste de s\'election de variables en grande dimension. Le nombre d'observations dans 
les \'etudes \'epid\'emiologiques \'etant faible, nous proposons une  validation crois\'ee  \`a deux niveaux 
pour \'eviter le risque de sur apprentissage dans la phase de s\'election de variables. Les estimateurs Lasso \'etant biais\'es et la variable d'int\'er\^et  qu'est le
nombre d'anoph\`eles \`a pr\'edire \'etant discret, nous utilisons un mod\`ele GLM pour d\'ebiaiser les variables s\'electionn\'ees par le Lasso et faire de la  pr\'ediction. 
Les r\'esultats montrent que quelques  variables climatiques et environnementales  
seulement sont des facteurs principaux li\'es au risque d'exposition au paludisme.
\smallskip

{\bf Mots-cl\'es.} Lasso, validation crois\'ee, s\'election de variables, pr\'ediction.
\bigskip\bigskip

{\bf Abstract.} In this study, we propose an automatic learning method for variables selection based on  Lasso in epidemiology
context.  One of the aim of this approach is to   overcome  the pretreatment of experts in medicine and epidemiology on collected data.
These pretreatment consist in recoding some variables and to choose some interactions based on expertise.
The approach proposed uses all available explanatory variables without  treatment and 
 generate automatically   all interactions between them. This lead to high dimension. We use Lasso, one of the robust methods of variable selection in high
 dimension. To avoid over fitting a  two levels cross-validation is used. Because the 
 target variable is account variable  and the lasso estimators are biased, 
 variables selected by lasso  are debiased by a GLM  and used to predict the distribution of the main vector of malaria which is Anopheles.
Results show that only few climatic and environmental variables are the mains factors associated to the malaria risk exposure.

\smallskip



\section{Introduction}
Le paludisme est un probl\`eme de sant\'e publique en Afrique surtout 
dans la zone sub-saharienne. Il constitue la premi\`ere cause de mortalit\'e 
pour  des enfants de moins de cinq ans et frappe 
essentiellement les couches les plus vuln\'erables de la population: les femmes enceinte et les nouveau-n\'es. 
Des \'etudes de cohorte ont \'et\'e conduites  
dans les zones end\'emiques pour \'etudier la mise en place et l'\'evolution du syst\`eme 
immunitaire du nouveau-n\'e face \`a cette maladie. 
Ces \'etudes ont aussi pour objectif d'\'etudier les d\'eterminants  li\'es \`a l'apparition des premi\`eres infections palustres 
chez le nouveau-n\'e. Certaines \'etudes ont montr\'e que la distribution du principal vecteur du paludisme qu'est l'anoph\`ele
ainsi que le risque d'exposition au paludisme pr\'esentent des d\'ependances  \`a la fois spatiales et  temporelles 
et non homog\`enes  \`a une petite \'echelle (niveau maison) [2]. Dans l'analyse et le traitement des donn\'ees issues de ces enqu\^etes,
les experts op\`erent des pr\'etraitements qui consistent entre autre \`a recoder certaines variables en classes et \`a choisir 
manuellement des interactions  de fa\c{c}on experte entre les variables explicatives.
Ils utilisent ensuite des m\'ethodes classiques de type \textit{forward, backward} pour la s\'election de variables [8].
L'objectif principal de ce travail est de s'affranchir de la phase de pr\'etraitement des experts m\'edicaux qui 
co\^ute en temps et qui pr\'esente un risque et de construire de fa\c{c}on automatique un mod\`ele pr\'edictif utilisant toutes les variables 
ainsi que toutes les interactions entres ces variables. Ce nombre \'elev\'e de variables nous conduit 
en grande dimension. Nous utilisons le   Lasso, une m\'ethode r\'egularisante qui fait \`a la fois de la s\'election est de l'estimation
et qui est robuste pour la s\'election de variables en grande dimension. 
 Dans les enqu\^etes \'epid\'emiologiques, les observations sont peu nombreuses.  Dans la s\'election de variables,  nous proposons  une validation crois\'ee \`a deux niveaux pour \'eviter  le risque
de sur apprentissage [7]. La variable d'int\'er\^et est le risque d'exposition au paludisme, qui revient au 
nombre d'anoph\`eles collect\'es dans les maisons donc discr\`ete alors nous utilisons  un mod\`ele simple de type  GLM avec un lien poisson. 
Ainsi  le GLM-Lasso permet de faire la s\'election de variables et le  GLM permet de d\'ebiaiser   les coefficients des variables 
s\'electionn\'ees par le Lasso pour la pr\'ediction.
 Les r\'esultats obtenus seront compar\'es \`a ceux de la m\'ethode de r\'ef\'erence (B-GLM) bas\'ee 
l'intervention des experts [2]. Ces r\'esultats montrent que quelques  variables climatiques et environnementales  
sont les facteurs principaux li\'es au risque d'exposition au paludisme.
\section{M\'ethodologie}
\subsection{Collecte des donn\'ees et variables utilis\'ees}
Les donn\'ees utilis\'ees dans ce travail proviennent d'une enqu\^ete \'epid\'emiologique conduite entre juillet 2007 et juillet 
2009 dans la commune de Tori-Bossito au B\'enin. Les donn\'ees sont de deux types : climatiques et environnementales (saison, quantit\'e de pluie, type 
de v\'eg\'etation, type de sol, etc), et des donn\'ees entomologiques (nombre de moustiques, nombre d'anoph\`eles infect\'es ou non.).
\subsection{Mod\`ele d'\'etude}
Le GLM-Lasso consiste  \`a p\'enaliser la log-vraisemblance du GLM en ajoutant une p\'enalit\'e $L_1$ [3,4,5].
Les coefficients des variables sont donn\'es par :
\begin{equation}\label{lasso}
 \hat{\beta} = Arg\max_{\beta}{\left[l_{GLM}(\beta| Y)+ \lambda  \sum_{i=1}^{p} {\beta_{i}}\right]}\quad \mbox{avec}\;\; \lambda \geq 0
\end{equation}
Le choix du param\`etre $\lambda$ se fait en minimisant le  score.
En pratique, l'\'equation (\ref{lasso}) n'a pas de solution num\'erique exacte. On utilise l'approximation de Laplace,
la m\'ethode de Newton-Raphson ou la m\'ethode du score de Fisher. Les coefficients du Lasso \'etant biais\'es, on utilise le GLM
pour les d\'ebiaiser et faire de la pr\'ediction. Sous forme matricielle le GLM se pr\'esente comme suit:
\begin{equation}
g[E(Y|\beta)]=X\beta 
\end{equation}
o\`u $(Y | \beta)$ suit une loi de Poisson de param\`etre  $E(Y | \beta)$, 
$n$\, est le nombre  observations, $ X$  la matrice   de dimensions $n \times \,(p+1)$
 des co-variables (environnementales et climatiques), 
$\beta$ est le vecteur de longueur $(p+1)$  des effects fixes y compris la constante, 
$Y$ est le vecteur des observations de la variable d'int\'er\^et. Ainsi
\begin{equation}
 \mathbb{P} ((Y=y_i|X=x)) = \frac{e^{(x\beta)^{y_i}}}{(y_i)!} \times e^{-e^{x\beta}} 
\end{equation}
Si on pose $Z_i=(Y=y_i|X=x) $ alors la vraisemblance des  $n$ observations peuvent \^etre d\'efinie comme:
 \begin{equation}
  L(Z_1, \ldots, Z_n) = \prod_ {i=1}^{n} \frac{e^{(x\beta)^{y_i}}}{(y_i)!} \times e^{-e^{x\beta}}
 \end{equation}
et la log-vraisemblance devient :
\begin{eqnarray}
  \mathcal{L}(Z_1, \ldots, Z_n) &= &\log\left( \prod_ {i=1}^{n} \frac{e^{(x\beta)^{y_i}}}{(y_i)!} \times e^{-e^{x\beta}}\right)
 \end{eqnarray}
 \begin{equation}
  \mathcal{L}(Z_1, \ldots, Z_n) =Cste + \sum^{n}_ {i=1} y_i(x\beta) - e^{(x\beta)}\quad \mbox{o\`u} \quad Cste =-\sum^{n}_ {i=1}\log((y_i)!)
 \end{equation}
\subsection{Algorithme LOLO-DCV}
L'algorithme  Leave one level out double cross-validation (LOLO-DCV) \'etudi\'e dans ce travail 
est bas\'e sur une validation crois\'ee statifi\'ee  \`a deux niveaux. Le deuxi\`eme niveau de validation crois\'ee permet d'\'eviter 
le risque de sur apprentissage dans la phase de s\'election de variables parce que le nombre d'observations n'est pas \'elev\'e. 
L'algorithme  se pr\'esente comme  d\'ecrit dans  (\ref{LOLO_DCV_Algorithme}).
 \begin{algorithm}[!ht]
 \caption{LOLO-DCV}
 \begin{enumerate}
  \item 
Les donn\'ees sont divis\'ees en $N$-blocs
 \item A chaque \'etape du premier niveau de la validation-crois\'ee
 \begin{enumerate}
\item Les blocs sont regroup\'es en deux parties : $E_A$ et $E_T$, 
$E_A$ : l'ensemble d'apprentissage qui contient les observations de 
 $(N-1)$-blocs,\\ $E_T$ :  l'ensemble de test, contenant les observations du dernier bloc. 
\item On met de c\^ot\'e  $E_T$
\item \label{CV1}  deuxi\`eme niveau de validation crois\'ee.
\begin{enumerate}
\item    On op\`ere  une validation-crois\'ee compl\`ete sur $E_A$ 
  \item  les deux param\`etres de r\'egularisation   $\lambda.min$ et $\lambda.1se$ sont r\'ecup\'er\'es.
  \item Les coefficients des variables actives  (variables \`a coefficient non nul) 
  associ\'es \`a ces deux param\`etres sont r\'ecup\'er\'es et d\'ebiais\'es.
    \item On utilise un mod\`ele GLM pour faire de la pr\'ediction sur  $E_T$
  \item La  pr\'esence $\mathcal{P}(X_i)$ de chaque variable est d\'etermin\'ee
  \end{enumerate}
\end{enumerate}
\item l'\'etape (\ref{CV1}) est r\'ep\'et\'ee jusqu'\`a faire de la pr\'ediction pour toutes les observations.
\end{enumerate}
\label{LOLO_DCV_Algorithme}
\end{algorithm}
Il  est bas\'e sur le score de validation qui est la d\'eviance du mod\`ele d\'efinit comme :
\begin{equation}
  Score( \lambda_{i}) =Deviance(\lambda_i)  = 2\times \left( \mathcal{L}_{(sat)} - \mathcal{L}_{(\lambda_i)}\right)
  \label{Score_Equation}
\end{equation}

o\`u $\mathcal{L}_{(sat)}$ est la log-vraisemblance du mod\`ele complet qui ajuste parfaitement les donn\'ees,
et $\mathcal{L}_{(\lambda_i)}$ 
log-vraisemblance du mod\`ele consid\'er\'e.
\begin{equation}
 Score( \lambda_{max})=Deviance(NULL)  = 2\times \left( \mathcal{L}_{(sat)} - \mathcal{L}_{\lambda_{max}}\right)
\end{equation}
Le mod\`ele obtenu \`a $\lambda=\lambda_{max}$ on obtient mod\`ele nul (le mod\`ele contenant uniquement l'intercept).
En posant 
\begin{equation}
 R= 1-\frac{Score( \lambda_{i})}{Score( \lambda_{max})}=1-r
\end{equation}
on a : $Deviance(\lambda_i) = (1-R)\times Score( \lambda_{max}) $.
On sait que $\mathcal{L}_{(sat)} = 0$ et ainsi $r$ devient le rapport de vraisemblance entre
le mod\`ele consid\'er\'e et le mod\`ele nul.

La valeur optimale  $\lambda.min$ de $\lambda$ est celle qui minimise la fonction $Score(.)$. 
\begin{equation}
 \lambda.min= Arg\min_{\lambda_{i}} [Score(\lambda_{i})]   
\end{equation}
La valeur $\lambda.1se$ est  telle que d\'efinie 
par T. Hastie et $al$ qui minimise le score plus sa d\'eviation standard [5].
Pour  $\lambda.min$ et $\lambda.1se$,  l'algorithme d\'etermine  les variables les plus fr\'equentes (Var$\_$freq), 
variables qui apparaissent un certain nombre de fois au premier niveau de la validation-crois\'ee
selon un seuil fix\'e. Ces sous ensembles de variables fr\'equentes  sont utilis\'ee pour  la pr\'ediction via un GLM.
\subsection{Pouvoir Pr\'edictif et crit\`ere de qualit\'e}
Les crit\`eres de qualit\'e utilis\'es pour la s\'election sont: La d\'eviance d\'efinie plus haut, la d\'eviance pond\'er\'ee $W.Deviance$
d\'efinie par : 
\begin{equation}
 W.Deviance(\lambda_i) = \frac{\frac{1}{w_i}\times Deviance(\lambda_i) }{\sum_i \frac{1}{w_i}}
\end{equation}
 o\`u  le nombre d'observations de l'ensemble d'apprentissage \\et le Pouvoir pr\'edictif $P_a$   d\'efini par:
\begin{displaymath}     
      \left\{
      \begin{array}{l}
   P_a(\hat Y_i) = 1 \quad \mbox{si}\; \;\; -0.5 \leq Y_i - \hat Y_i \leq 0.5 \quad \cr
   P_a(\hat Y_i) = 0 \quad \mbox{sinon}.
              \end{array}
                         \right.
      \end{displaymath}
      o\`u    $ \hat Y_i$ est la pr\'ediction pour chaque observation $Y_i$.     
\section{R\'esultats et Conclusion}
\begin{table}
\caption{
\bf{Crit\`eres de qualit\'e pour  B-GLM}} \label{Summary_orig}
\begin{center}
\begin{tabular}{|l|r|r|r|r|r|}
\hline
M\'ethode & Deviance & W.Deviance & Pouvoir pr\'edictif (\%)\\
\hline
B-GLM & 3101.68 & 3101.49& 73.53\\
\hline
\end{tabular}
\end{center}
\end{table}
\begin{table}
\caption{
\bf{Crit\`eres de qualit\'e pour LOLO-DCV}  }
\label{selection_variables_originales_avec_village_table}
\begin{center}
\begin{tabular}{|l|r|r|r|}
\hline
 M\'ethode & Deviance & W.Deviance & Pouvoir pr\'edictif (\%)\\
\hline
LOLO DCV lamba\_min & 5573.98 & 5573.67 & 78.76\\
\hline
LOLO DCV lamba\_1se & 5573.98 & 5573.67 & 78.76\\
\hline
Var freq lamba\_min & 2860.75 & 2860.59 & 75.00\\
\hline
Var freq lamba\_1se & 3259.69 & 3259.53 & 76.80\\
\hline
\end{tabular}

\end{center}
\end{table}
Le meilleur sous-ensemble optimal de variables pour chaque m\'ethode est :\\
 \textbf{B-GLM}: 
  La saison, le nombre de jours de pluie, la quantit\'e moyenne de pluie, l'utilisation de r\'epulsif, la v\'eg\'etation,
  l'interaction entre saison et la v\'eg\'etation.\\
\textbf{LOLO-DCV}: 
 La saison et l'interaction entre le nombre de jours de pluie et le village.
 Les r\'esultats  des tables (\ref{Summary_orig}, \ref{selection_variables_originales_avec_village_table}) 
 montrent que les meilleurs pr\'edictions sont obtenues par LOLO-DCV et le sous ensemble optimale pour la pr\'ediction 
 de LOLO-DCV est plus parcimonieux que celui obtenu par la m\'ethode (B-GML). Ces r\'esultats montrent  que la machine peut remplacer les experts
pour la s\'election de variables et am\'eliorer leurs r\'esultats.

\section*{Bibliographie}
\noindent [1] De Brabanter J., Pelckmans K., Suykens J.A.K., Vandewalle J. , (2002) 
Robust cross-validation score function for LS-SVM non-linear function estimation,
\textit{: Int. Conference on Artificial Neural Networks - ICANN}, pp 713-719.

\noindent [2] G. Cottrell,  B. Kouway\`e and $al$, (2012)  Modeling the 
Influence of Local Environmental Factors on Malaria 
Transmission in Benin and Its Implications for Cohort Study, \textit{PlosOne}, 7(8).

\noindent [3] B. Efron, T. Hastie, I. Johnstone, and R. Tibshirani. (2004),Least angle regression, \textit{The Annals of statistics.}
, 32:407-499.

\noindent [4]  R. Tibshirani, (1996):  Regression shrinkage and selection via the lasso, \textit{Journal of 
the Royal Statistical Society Serie B (Methodological)}, 58:267-288.

\noindent [5] H. Zou and T. Hastie, (2005): Regularization and variable selection via the elastic net, 
\textit{Journal of the Royal Statistical Society. Serie B}, 67:301-320.

\noindent [6]  M. Osborne,  B. Presnell and B.  Turlach, (2000): A new approach to 
variable selection in least squares problems, \textit{ IMA Journal of Numerocal Analysis},
20:389-403.

\noindent [7] A. Y. Ng, (1997), Preventing "Overfitting" of Cross-Validation Data,
\textit{International Conference on Machine Learning}, pp 245-253.

\noindent [8] I. Guyon,  B. Presnell and B.  Turlach, 
An Introduction to Variable and Feature Selection, 
3:1157-1182, Journal of Machine Learning Research, 2003.
\noindent [9]  B Gianluca. : Structural feature selection for wrapper methods, editor, 
\emph{proceedings of the $13^{th}$
European Symposium on Artificial Neural Networks} ({ESANN} 2005),
d-side pub., pages 405-410, April 27-29, Bruges (Belgium), 2005.
\end{document}